\documentclass[information,article,accept,pdftex,moreauthors]{Definitions/mdpi} 
\firstpage{1} 
\pubvolume{1}
\issuenum{1}
\articlenumber{0}
\pubyear{2023}
\copyrightyear{2023}
\externaleditor{Academic Editor: Alessandra Lumini}
\datereceived{31 July 2023} 
\daterevised{15 September 2023} 
\dateaccepted{20 September 2023} 
\datepublished{ } 
\hreflink{https://doi.org/} 
\usepackage{amsmath,amssymb,amsfonts}
\DeclareMathOperator*{\argmax}{arg\,max}

\usepackage{algpseudocode}
\usepackage{algorithm}

\newcommand{\cut}[1]{ }
\usepackage{enumitem}

\newcommand{\kk}[1]{}
\usepackage[none]{hyphenat}

\Title{AdvRain: Adversarial Raindrops to Attack Camera-Based Smart Vision Systems}

\TitleCitation{AdvRain: Adversarial Raindrops to Attack Camera-Based Smart Vision Systems}

\Author{Amira Guesmi 
 *, Muhammad Abdullah Hanif and Muhammad Shafique} 

\AuthorNames{Amira Guesmi, Muhammad Abdullah Hanif, and Muhammad Shafique}

\AuthorCitation{Guesmi, A.; Hanif, M.A.; Shafique, M.}

\address[1]{%
eBrain Lab, Division of Engineering, New York University Abu Dhabi (NYUAD), Abu Dhabi 129188, United Arab Emirates; 
mh6117@nyu.edu 
 (M.A.H.); ms12713@nyu.edu (M.S.)\\
}

\corres{\hangafter=1 \hangindent=1.05em \hspace{-0.82em} Correspondence: ag9321@nyu.edu}

\abstract{Vision-based perception modules are increasingly deployed in many applications, especially autonomous vehicles and intelligent robots. These modules are being used to acquire information about the surroundings and identify obstacles. Hence, accurate detection and classification are essential to reach appropriate decisions and take appropriate and safe actions at all times.
Current studies have demonstrated that ``printed adversarial attacks'', known as physical adversarial attacks, can successfully mislead perception models such as object detectors and image classifiers. However, most of these physical attacks are based on noticeable and eye-catching patterns for generated perturbations making them identifiable/detectable by the human eye, in-field tests, or in test drives. 
In this paper, we propose a camera-based inconspicuous adversarial attack (\textbf{AdvRain}) 
 capable of fooling camera-based perception systems over all objects of the same class. Unlike mask-based FakeWeather attacks that require access to the underlying computing hardware or image memory, our attack is based on emulating the effects of a natural weather condition (i.e., Raindrops) that can be printed on a translucent sticker, which is externally placed over the lens of a camera whenever an adversary plans to trigger an attack. Note, such perturbations are still inconspicuous in real-world deployments and their presence goes unnoticed due to their association with a natural phenomenon.  
To accomplish this, we develop an iterative process based on performing a random search aiming to identify critical positions to make sure that the performed transformation is adversarial for a target classifier. Our transformation is based on blurring predefined parts of the captured image corresponding to the areas covered by the raindrop.
We achieve a drop in average model accuracy of more than $45\%$ and $40\%$ on VGG19 for ImageNet dataset 
 and Resnet34 for Caltech-101 dataset, respectively, using only $20$ raindrops. 
}


\begin{document}




\section{Introduction}
The revolutionary emergence of deep learning (DL) has shown a profound impact across diverse sectors, particularly in the realm of autonomous driving \cite{al2017deep} 
. Prominent players in the automotive industry, such as Google, Audi, BMW, and~Tesla, are actively harnessing this cutting-edge technology in conjunction with cost-effective cameras to develop autonomous vehicles (AVs). These AVs are equipped with state-of-the-art vision-based perception modules, empowering them to navigate real-life scenarios even under high-pressure circumstances, make informed decisions, and~execute safe and appropriate~actions.

Consequently, the~demand for autonomous vehicles has soared, leading to substantial growth in the AV market. Strategic Market Research (SMR) predicts that the autonomous vehicle market will achieve an astonishing valuation of \$196.97 billion by 2030, showcasing an impressive compound annual growth rate (CAGR) of 25.7\% (ACMS). The~integration of DL-powered vision-based perception modules has undeniably accelerated the progress of autonomous driving technology, heralding a transformative era in the automotive industry. With~the increasing prevalence of AVs, their potential impact on road safety, transportation efficiency, and~overall user experience remains a subject of great interest to consumers, researchers, and~investors~alike.

However, despite the significant advancements in deep learning models, they are not immune to adversarial attacks, which can pose serious threats to their integrity and reliability. Adversarial attacks involve manipulating the input of a deep learning classifier by introducing carefully crafted perturbations, strategically chosen by malicious actors, to~force the classifier into producing incorrect outputs. Such vulnerabilities can be exploited by attackers to compromise the security and integrity of the system, potentially endangering the safety of individuals interacting with it. For~instance, a~malicious actor could add adversarial noise to a stop sign, causing an autonomous vehicle to misclassify it as a speed limit sign~\cite{b0,fgsm}. This kind of misclassification could lead to dangerous consequences, including accidents and loss of life. Notably, adversarial examples have been shown to be effective in real-world conditions~\cite{phy9}. Even when printed out, an~image specifically crafted to be adversarial can retain its adversarial properties under different lighting conditions and~orientations.

Therefore, it becomes crucial to understand and mitigate these adversarial attacks to ensure the development of safe and trustworthy intelligent systems. Taking measures to defend against such attacks is imperative for maintaining the reliability and security of deep learning models, particularly in critical applications such as autonomous vehicles, robotics, and~other intelligent systems that interact with~people.

Adversarial attacks can broadly be categorized into two types: \textit{Digital Attacks 
} and \textit{Physical Attacks}, each distinguished by its unique form of attack~\cite{fgsm, CW, phy9}. In~a \textit{Digital Attack}, the~adversary introduces imperceptible perturbations to the digital input image, specifically tailored to deceive a given deep neural network (DNN) model. These perturbations are carefully optimized to remain unnoticed by human observers. During~the generation process, the~attacker works within a predefined noise budget, ensuring that the perturbations do not exceed a certain magnitude to maintain imperceptibility.
In contrast, \textit{Physical Attacks} involve crafting adversarial perturbations that can be translated into the physical world. These physical perturbations are then deployed in the scene captured by the victim DNN model. Unlike digital attacks, physical attacks are not bound by noise magnitude constraints. Instead, they are primarily constrained by location and printability factors, aiming to generate perturbations that can be effectively printed and placed in real-world settings without arousing~suspicion.

The primary objective of an adversarial attack and its relevance in real-world scenarios is to remain inconspicuous, appearing common and plausible rather than overtly hostile. Many previous works in developing adversarial patches for image classification have focused mainly on maximizing attack performance and enhancing the strength of adversarial noise. However, this approach often results in conspicuous patches that are easily recognizable by human observers.
Another line of research has aimed to improve the stealthiness of the added perturbations by making them blend seamlessly into natural styles that appear legitimate to human observers. Examples include camouflaging the perturbations as color films~\cite{Zhong2022}, shadows~\cite{Hu2022}, or~laser beams~\cite{Duan2020}, among~others.

\begin{figure}[H]
\hspace{-6pt}\includegraphics[width=0.7\columnwidth]{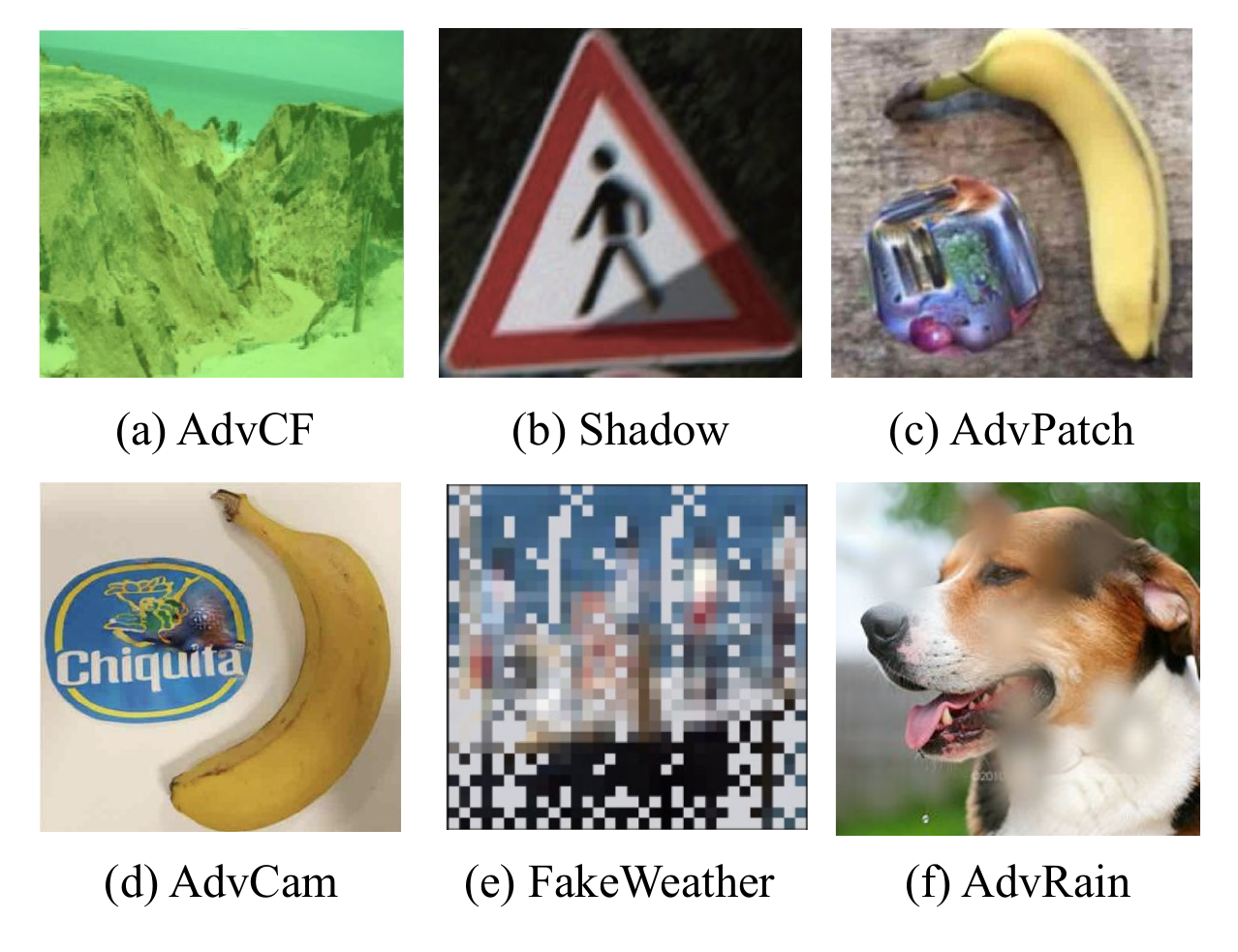}
\caption{AdvRain vs. existing physical-world attacks: (\textbf{a}) AdvCF~\cite{Hu2022}, (\textbf{b}) Shadow~\cite{Zhong2022},\linebreak (\textbf{c}) advPatch~\cite{googleap}, (\textbf{d}) AdvCam~\cite{Duan2020}, (\textbf{e}) FakeWeather~\cite{fakeweather}, and~(\textbf{f}) AdvRain. }
\label{sota_attacks}
\end{figure}

In Figure 
 \ref{sota_attacks}, we provide a visual comparison of AdvRain with existing physical attacks. While all the adversarial examples in Figure~\ref{sota_attacks} successfully attack deep neural networks (DNNs), AdvRain stands out in its ability to generate adversarial perturbations with natural blurring marks (emulating a similar phenomenon to the actual rain), unlike the conspicuous pattern generated by AdvPatch or the unrealistic patterns generated by FakeWeather~\cite{fakeweather}. This showcases the effectiveness of AdvRain in creating adversarial perturbations that blend in with the surrounding environment, making them difficult for human observers to~detect.

\begin{figure}[H]
\hspace{-7.5pt}\includegraphics[width=0.8\columnwidth]{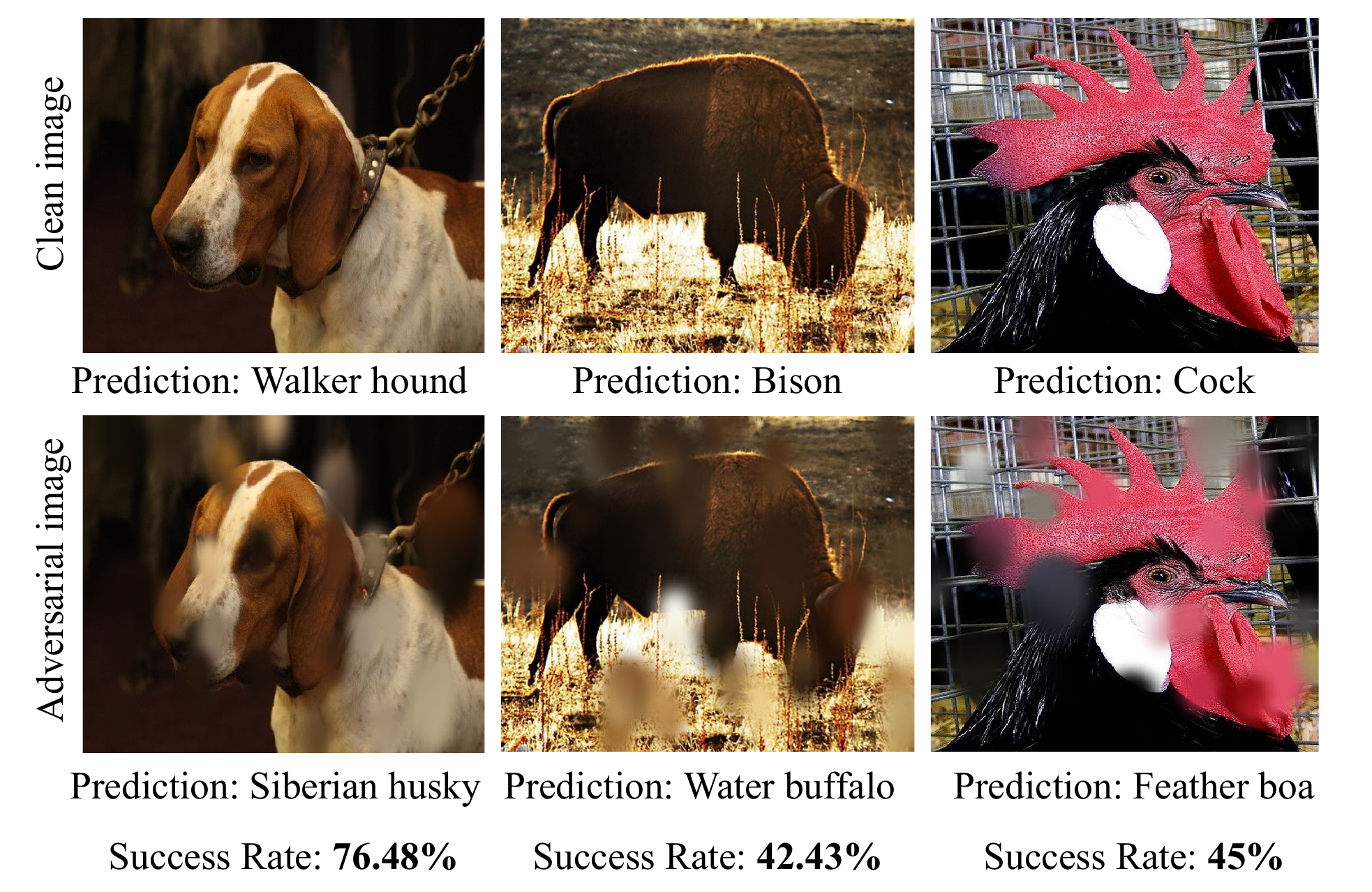}
    \caption{Adversarial 
 examples generated by \textbf{AdvRain} and their corresponding attack success rate when using only 10 raindrops.}
\label{advrain}
\end{figure}

In this paper, we present a novel technique aimed at deceiving a given DNN model by introducing a subtle perturbation that causes misclassification of all objects belonging to a specific class. Our approach involves creating an adversarial camera sticker, designed to be attached to the camera's lens. 
 This sticker features a carefully crafted pattern of raindrops, which, when perceived by the camera, leads to misclassification of the captured images (See Figure \ref{advrain}). The~patterns created by the raindrops appear as water drops in the camera image, making them inconspicuous to human~observers.
 
Unlike previous adversarial attacks that typically operate at the pixel-level granularity of the images, our main challenge lies in the limited space of feasible perturbations that can be introduced using this camera sticker model. The~physical optics of the camera constrain us, resulting in the production of blurry dots as the primary perturbation. These blurry dots lack the high-frequency patterns commonly found in traditional adversarial attacks.
Our research addresses the unique challenges posed by physical adversarial attacks, where the goal is to exploit the limitations and characteristics of the camera's optics to achieve stealthiness. By~developing this technique, we contribute to the understanding of physical adversarial attacks and provide insights into developing robust defense mechanisms for intelligent systems in real-world~scenarios.

\begin{figure}[H]
\hspace{-9pt}\includegraphics[width=\columnwidth]{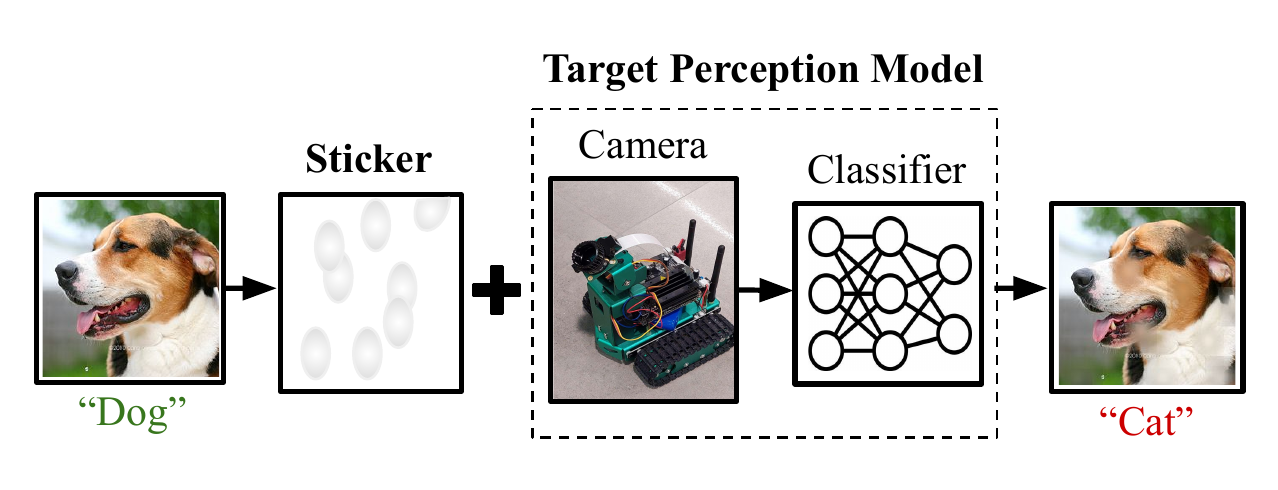}
    \caption{\textbf{Attack threat model: 
} The 
 generated pattern is printed on a translucent sticker placed over the lens of the camera. Hence, any captured image will contain the adversarial dots resulting in an inconspicuous, natural-looking adversarial image that fools the target model and the human eye.}
\label{threat}
\end{figure}

In this paper, we present a novel technique for crafting adversarial perturbations using a random search optimization method guided by Grad-CAM~\cite{grad}. By~using Grad-CAM, 
 we are able to identify critical positions that are likely to result in higher attack success rates. This enables us to focus on exploring a smaller set of positions during the random search, ultimately determining the best raindrop positions that ensure the highest effectiveness of our AdvRain attack. 
Our objective is to create perturbations that can be introduced into the visual path between the camera and the object, while keeping the object itself unaltered (As presented in Figure \ref{threat}). By~leveraging grad-cam, we can identify critical positions in the image that significantly influence the decision-making process of the deep learning model. This information guides our search optimization method, helping us generate adversarial perturbations at those influential positions. As~a result, the~perturbations effectively deceive the target classifier without directly modifying the object. The~advantage of our approach lies in its ability to subtly manipulate the visual information captured by the camera. The~crafted adversarial perturbations blend seamlessly into the scene, remaining inconspicuous to human observers. Simultaneously, they have a substantial impact on the model's decision, leading to misclassification of the object.

\vspace{-15pt}

\begin{figure}[H]
\includegraphics[width=1\columnwidth]{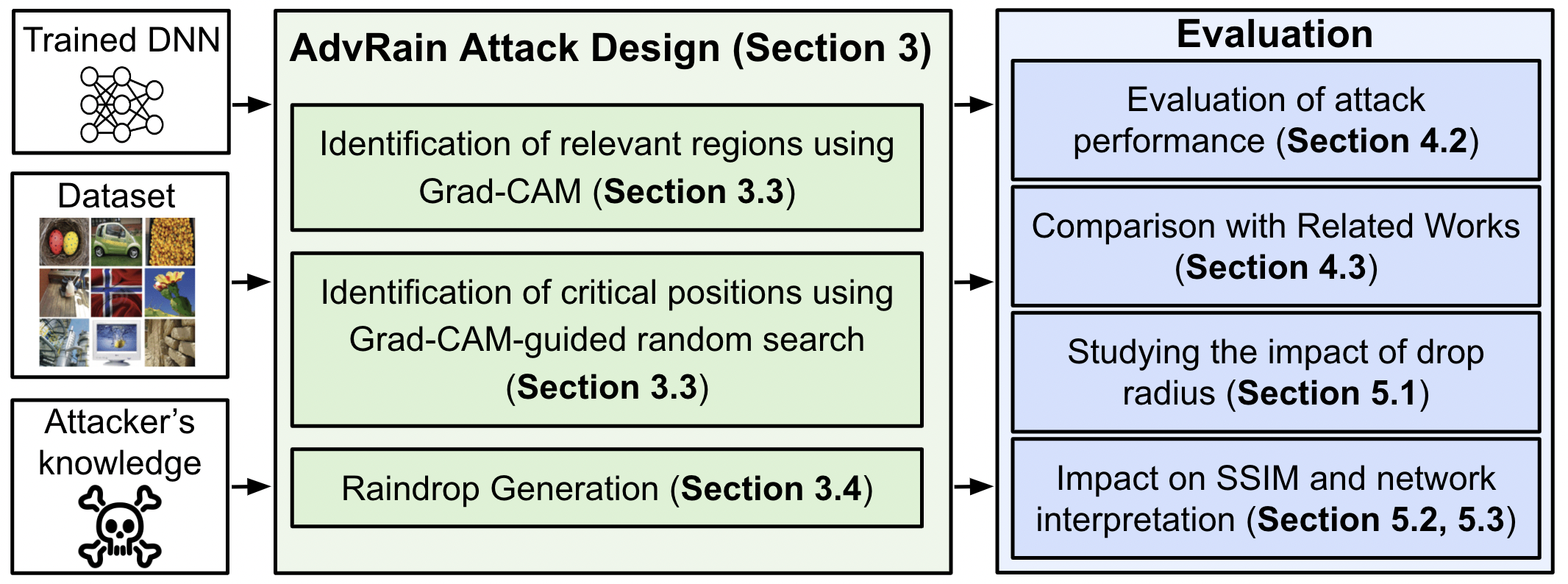}
    \caption{Overview 
 of our Novel Contributions. 
}
\label{contribution}
\end{figure}

An overview of our novel contributions is shown in Figure~\ref{contribution}.

\textbf{In summary, the~\textit{contributions} of this work are:} 

\begin{itemize}
    \item We propose a novel technique that utilizes a random search optimization method guided by grad-cam to craft adversarial perturbations. These perturbations are introduced into the visual path between the camera and the object without altering the appearance of the object~itself.

    \item The adversarial perturbations are designed to resemble a natural phenomenon, specifically raindrops, resulting in an inconspicuous pattern. These patterns are printed on a translucent sticker and affixed to the camera lens, making them difficult to~detect.

    
    
    \item The proposed adversarial sticker applies the same perturbation to all images belonging to the same class, making it a universal attack against the target class.
    \item Our experiments demonstrate the potency of the AdvRain attack, achieving a significant decrease of over 61\% in accuracy for VGG-19 on ImageNet and 57\% for Resnet34 on Caltech-101 compared to 37\% and 40\% (for the same structural similarity index (SSIM)) when using FakeWeather 
~\cite{fakeweather}.
    \item We study the impact of blurring specific parts of the image, introducing low-frequency patterns, on~model interpretability. This provides valuable insights into the behavior of the deep learning models under the proposed adversarial attack.
\end{itemize}

Paper Organization: The structure of the remaining article is as follows. Section~\ref{sec:related} offers a comprehensive overview of related work, setting the context for our proposed approach. In~Section~\ref{sec:proposed}, we present the methodology for generating the adversarial sticker, outlining the key components of our AdvRain attack. Next, in~Section~\ref{sec:experiment}, we conduct a detailed evaluation of the proposed attack. We compare AdvRain against FakeWeather and Natural rain, examining the attack's potency and the visual similarity of the generated adversarial examples. Additionally, we examine the attack's potency and the visual similarity of the generated adversarial examples. In~Section~\ref{sec:discussion}, we thoroughly discuss the findings and implications of our experiments. Additionally, we explore the impact of drop radius, assess the perturbation's effect on SSIM, and~investigate its influence on network interpretation.
Finally, in~Section~\ref{sec:conclusion}, we provide a concise summary and conclusion of our study, highlighting the significance of AdvRain as a stealthy and effective approach to adversarial attacks in camera-based vision~systems.

\section{ Background \& Related~Work}
\label{sec:related}
\unskip

\subsection{Camera-Based Vision~Systems}


Environment perception has emerged as a crucial application, driving significant efforts in both the industry and research community. The~focus on developing powerful deep learning (DL)-based solutions has been particularly evident in applications such as autonomous robots and intelligent transportation systems. Designing reliable recognition systems is among~the primary challenges in achieving robust environment perception. In~this context, automotive cameras play a pivotal role, along with associated perception models such as object detectors and image classifiers, forming the foundation for vision-based perception modules. These modules are instrumental in gathering essential information about the environment, aiding autonomous vehicles (AVs) in making critical decisions for safe~driving.

The pursuit of dependable recognition systems represents a major hurdle in establishing high-performance environment perception. The~accuracy and reliability of such systems are critical for the safe operation of AVs and ensuring their successful integration into real-world scenarios. However, it has been demonstrated that these recognition systems are susceptible to adversarial attacks, which can undermine their integrity and pose potential risks to AVs and their passengers. In~light of these challenges, addressing the vulnerabilities of DL-based environment perception systems to adversarial attacks becomes paramount. Research and development efforts must focus on building robust defense mechanisms to fortify these systems against potential threats, enabling the safe and trustworthy deployment of AVs and intelligent transportation systems in the~future.
 
Adversarial attacks can be categorized into digital and physical attacks.
\subsection{Digital Adversarial~Attacks}
In scenarios involving digital attacks, the~attacker has the flexibility to manipulate the input image of a victim deep neural network (DNN) at the pixel level. These attacks assume that the attacker has access to the DNN's input system, such as a camera or other means of providing input data. The~concept of adversarial examples, where a small, imperceptible noise is injected to shift the model's prediction towards the wrong class, was first introduced by Szegedy~et~al.~\cite{Szegedy}.

Over time, various algorithms for creating adversarial examples have been developed, leading to the advancement of digital attacks. Some notable digital attacks include Carlini and Wagner's attack (CW) \cite{CW}, Fast Gradient Sign Method (FGSM) \cite{fgsm}, Basic Iterative Method (BIM) \cite{BA}, local search attacks~\cite{localsearch}, and~HopSkipJump attack \mbox{(HSJ) \cite{HSJ}}, among~others. However, in~a realistic threat model, we may assume that the attacker has control over the system's external environment or external objects, rather than having access to the system's internal sensors and data pipelines. This scenario represents physical attacks, where the attacker crafts adversarial perturbations that can be introduced into the physical world to deceive the~DNN.

In the following section, we will explore some state-of-the-art physical attacks on image classification. These attacks are designed to exploit the vulnerabilities of camera-based vision systems and demonstrate the potential risks of adversarial manipulation in real-world scenarios. Understanding and mitigating such physical attacks are crucial for building robust and secure intelligent systems, particularly in safety-critical applications such as autonomous vehicles and surveillance systems.
\subsection{Physical Adversarial~Attacks}

A physical attack involves adding perturbations in the physical space to deceive the target model. The~process of crafting a physical perturbation typically involves two main steps. First, the~adversary generates an adversarial perturbation in the digital space. Then, the~goal is to reproduce this perturbation in the physical space, where it can be perceived by sensors such as cameras and radars, effectively fooling the target~model.

Existing methods for adding adversarial perturbations in different locations can be categorized into four main groups:
\noindent\textit{Attack by Directly Modifying the Targeted Object}: 
 In this approach, the~attacker directly modifies the targeted object to introduce the adversarial perturbation. For~example, adversarial clothing has been proposed, where clothing patterns are designed to confuse object detectors~\cite{Hu21, guesmi2023advart,guesmi2023dap}. \textcolor{black}{Hu~et~al.~\cite{Hu21} leverage pretrained GAN models to generate realistic/naturalistic images that can be printed on t-shirts and are capable of hiding the person wearing them. Guesmi~et~al.~\cite{guesmi2023advart} proposed replacing the GAN with a semantic constraint based on adding a similarity term to the loss function and, in doing so, directly manipulating the pixels of the image. This results in a higher flexibility to incorporate multiple transformations. 
}

\textcolor{black}{\textit{Attack by Modifying the Background} 
:
Adversarial patches represent a specific category of adversarial perturbations designed to manipulate localized regions within an image, aiming to mislead classification models. These attacks leverage the model's sensitivity to local alterations, intending to introduce subtle changes that have a substantial impact on the model's predictions. By~exploiting the model's reliance on specific image features or patterns, adversaries can create patches that trick the model into misclassifying the image or perceiving it differently from its actual content. An~example of a practical attack for real-world scenarios is AdvPatch~\cite{googleap}. 
This attack creates universal patches that can be applied anywhere. Additionally, the~attack incorporates Expectation over Transformation (EOT) \cite{eot} to enhance the robustness of the adversarial patch.
The AdvCam technique~\cite{Duan2020} presents an innovative approach to image perturbation, operating within the style space. This method combines principles from neural style transfer and adversarial attacks to craft adversarial perturbations that seamlessly blend into an image's visual style. For~instance, AdvCam can introduce perturbations such as rust-like spots on a stop sign, making them appear natural and inconspicuous within their surroundings. 
}

\textit{Modifying the Camera}: This method involves modifying the camera itself to introduce the adversarial perturbation. One approach is to leverage the Rolling Shutter Effect, where the timing of capturing different parts of the image is manipulated to create perturbations~\cite{rolling, slmattack}. 

\textcolor{black}{\textit{Modifying the Medium Between the Camera and the Object}: This category includes attacks that modify the medium between the camera and the object. For~instance, light-based attacks use external light sources to create perturbations that are captured by the camera and mislead the target model~\cite{optical, Zhong2022}. The~Object Physical Adversarial Device \mbox{(OPAD) \cite{optical}} employs structured lighting methods to alter the appearance of a targeted object. This attack system is composed of a cost-effective projector, a~camera, and~a computer, enabling the manipulation of real-world objects in a single shot.
Zhong~et~al.~\cite{Zhong2022} harness the natural phenomenon of shadows to create adversarial examples. This method is designed to be practical in both digital and physical contexts. Unlike traditional gradient-based optimization algorithms, it employs optimization strategies grounded in particle swarm optimization (PSO) \cite{488968}. The~researchers conducted extensive assessments in both simulated and real-world scenarios, revealing the potential threat posed by shadows as a viable avenue for attacks.
However, it is important to note that these techniques may experience reduced effectiveness under varying lighting conditions.
The Adversarial Color Film (AdvCF) method, introduced by Zhang~et~al.~\cite{Hu2022}, utilizes a color film positioned between the camera lens and the subject of interest to enable effective physical adversarial attacks. By~adjusting the physical characteristics of the color film without altering the appearance of the target object, AdvCF aims to create adversarial perturbations that maintain their effectiveness in various lighting conditions, including both daytime and nighttime settings.
}

\textcolor{black}{
FakeWeather~\cite{fakeweather} attack aims to emulate the effects of various weather conditions, such as rain, snow, and~hail, on~camera lenses. This attack seeks to deceive computer vision systems, particularly those used in autonomous vehicles and other image-based applications, by~adding perturbations to the captured images that mimic the distortions caused by adverse weather. In~the FakeWeather attack, the~adversary designs specific masks or patterns that simulate the visual artifacts produced by different weather conditions. These masks are then applied to the camera's images, introducing distortions that can mislead image recognition models. The~goal is to make the images appear as if they were captured in inclement weather, potentially causing the models to make incorrect predictions or classifications. One limitation of the FakeWeather attack is that the generated noise or perturbations may have unrealistic and pixelated patterns, which could potentially be detected by more robust image recognition systems. Additionally, the~attack's effectiveness may be limited to specific scenarios and image sizes, as~it was initially tested on small images of 32 $\times$ 32 pixels from the CIFAR-10 dataset.
}

Additionally, researchers have explored techniques to create adversarial perturbations with natural styles to ensure stealthiness and legitimacy to human observers. Such approaches aim to make the perturbations appear as natural phenomena in the scene. 

\section{Proposed~Approach}
\label{sec:proposed}

In this section, we outline our methodology for designing the adversarial camera sticker, a~novel approach to creating inconspicuous adversarial perturbations for camera-based vision systems.
\subsection{Threat Model for Physical Camera Sticker~Attacks}
Traditionally, the~problem of generating an adversarial example is formulated as a constrained optimization (Equation (\ref{eq:adv})), given an original input image $x$ and a target classification model $ f(.) $: 
\begin{equation}
\label{eq:adv}
    \min_{\delta} \left\|\delta\right\|_p  \\
    s.t. f(x + \delta) \neq f(x)
\end{equation}
where 
 the objective is to find a minimal inconspicuous universal perturbation, $\delta$, 
 such that when added to an arbitrary input from a target input domain $D$, it will cause the underlying DNN-based model $f(.)$ to misclassify. Note that one cannot find a closed form solution for this optimization problem since the DNN-based model $f(.)$ is a non-convex machine learning model, i.e.,~a deep neural network. Therefore, Equation~(\ref{eq:adv}) is formulated as follows to numerically solve the problem using empirical approximation techniques:
\begin{equation}
\label{eq:formulation}
    \argmax_{\delta} \sum_{x \in \mathcal{D}} l(f(x+\delta), f(x))
\end{equation}
where $l$ is the DNN-based model loss function and $\mathcal{D} \subset D$ is the attacker’s classifier training dataset. To~solve this problem, existing optimization techniques (e.g., Adam~\cite{adam}) can be used. In~each iteration of the training the optimizer updates the adversarial noise $\delta$.

In contrast to attacks that operate at the pixel-level granularity of the images, our proposed threat model, the~physical camera sticker attacks, faces a significant challenge due to the limited space of feasible perturbations that can be introduced. The~optics of the camera impose constraints, resulting in the generation of only blurry dots as perturbations, lacking the high-frequency patterns typically found in traditional adversarial attacks. To~address this challenge, we propose an innovative approach to design the adversarial perturbation by approximating the effect of placing small dots, resembling raindrops, on~a sticker. When a small water drop is placed on the camera lens, it creates a translucent patch on the captured image. This patch represents the introduced low-frequency perturbations resulting from the optics of the camera lens. The~adversarial example is crafted as follows:
\begin{equation}
    x_{adv} = G(x,p,n,r)
\end{equation}
where $G$ is the raindrop generator, $p$ stands for the drops positions, $n$ for number of drops, $r$ for the drop radius.
Technically, considering a 2D image $x$ with $x(i,j)$ denoting the pixel at the $(i,j)$ location. 
Our aim is to find the best position candidate $p(i,j)$ for the $n$ raindrops, in~a way that the model wrongly classifies the generated adversarial example. 
\begin{equation}
        p(i,j)? ~  \\
    s.t.~ f(G(x,p(i,j),n,r)) \neq f(x)
\end{equation}

\subsection{Overview of the Proposed~Approach}
\begin{figure}[H]
\hspace{-15pt}\includegraphics[width=\columnwidth]{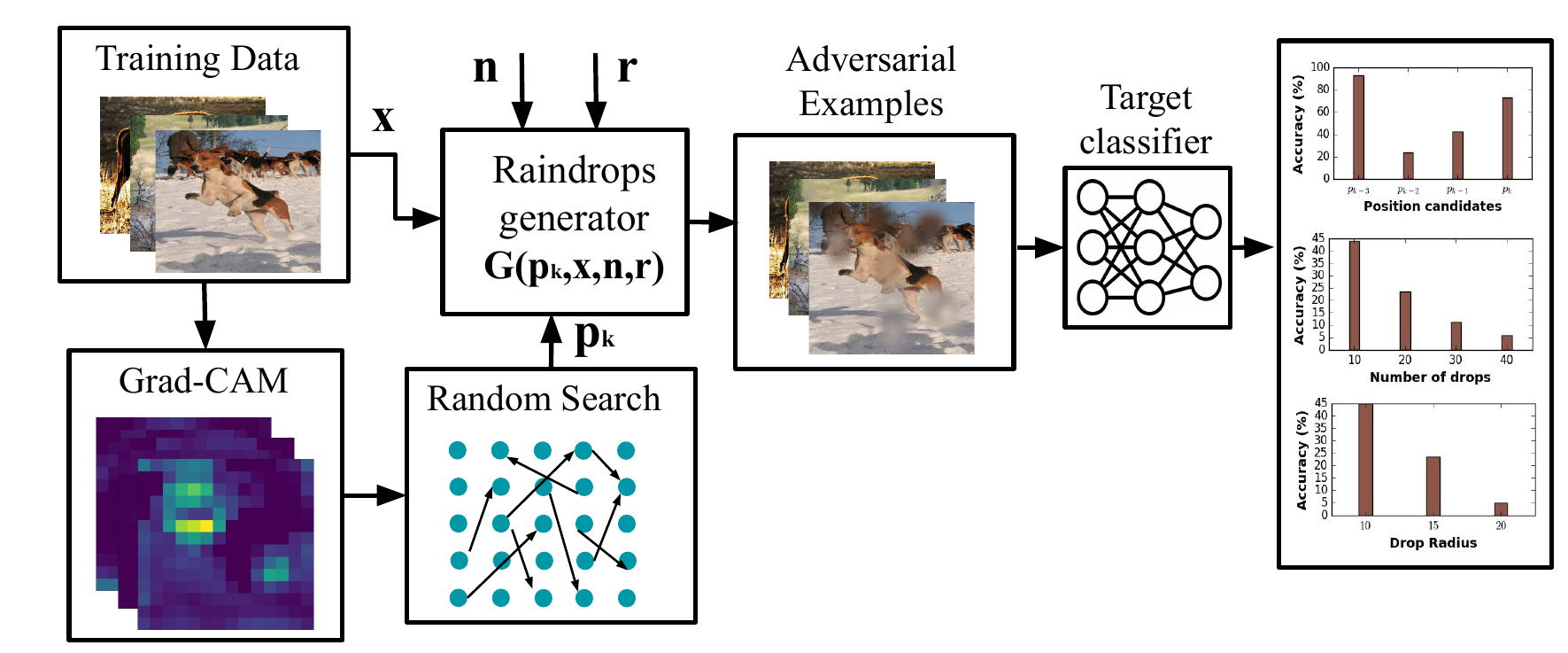}
\caption{Overview of the proposed approach; \textbf{Training phase:} 
 we employ a random search-based optimization method and leverage Grad-CAM to enhance the efficiency of the optimization process. By~using Grad-CAM, we are able to identify critical positions that are likely to result in higher attack success rates. This enables us to focus on exploring a smaller set of positions during the random search, ultimately determining the best raindrop positions that ensure the highest effectiveness of our AdvRain attack.
The raindrop generator takes as inputs the number of drops $n$, the~radius of the drop $r$, the~positions $p_k$ of the $n$ drop for the $k$ iteration, and~the input image $x$. The~output of the generator is the simulated raindrops added to the input image. This image is later on fed to the classifier to monitor its classification accuracy.}
\label{overview}
\end{figure}

\textls[-15]{Figure~\ref{overview} provides an overview of our proposed attack. The~main objective is to generate adversarial perturbations with patterns resembling the effect of natural weather events, particularly raindrops on the camera lens due to atmospheric conditions (i.e., Rain). To~achieve this, we craft patterns that simulate the appearance of raindrops on the camera lens.} 

\textcolor{black}{AdvRain utilizes a unique approach to create adversarial perturbations that mimic the effect of natural weather conditions, particularly rain, on~camera lenses. This approach can be broken down into several key steps: The process begins with the generation of a pattern that resembles the appearance of raindrops on a camera lens. These raindrops are a form of low-frequency perturbations that blur portions of the captured image. The~goal is to make these perturbations inconspicuous and blend them seamlessly with the visual characteristics of real raindrops. }

\textcolor{black}{
To optimize the pattern of raindrops for maximum effectiveness, a~random search guided by Grad-CAM (Gradient-weighted Class Activation Mapping) is employed. Grad-CAM helps identify important regions within the image that influence the model's decision. By~starting the optimization process from positions highlighted by Grad-CAM, the~search narrows down and explores a smaller set of potential perturbation patterns. Once a pattern is selected, it is applied to the input image as simulated raindrops. Each pixel within the area covered by a raindrop undergoes a transformation, specifically a Gaussian blur. This blurring effect mimics the distortion introduced by raindrops on a camera lens. In~addition to the blur, AdvRain also incorporates a fish-eye effect to simulate the curvature that raindrops can introduce to the camera's view. The~splashing of raindrops is simulated using collision detection, ensuring that the raindrops do not overlap in a physically implausible manner.  }

\textcolor{black}{
To create a natural raindrop appearance, AdvRain employs a combination of a circle and an oval shape for each raindrop. The~final shape represents the water droplet's surface, and~the blur applied to this shape adds to the realism. The~generated perturbation, resembling a collection of raindrops, is intended to be printed on a translucent sticker that can be affixed to the camera's lens. In~essence, AdvRain leverages a combination of random search optimization, Grad-CAM guidance, and~the transformation of carefully designed raindrop patterns to create physically inconspicuous adversarial perturbations. These perturbations, once applied to the camera lens, introduce low-frequency distortions that can deceive deep neural networks into misclassifying images. }
\vspace{-6pt}
\unskip

\subsection{Identify Critical~Positions}

The identification of critical positions is achieved using a Grad-CAM-guided Random Search Method, as~outlined in Algorithm \ref{algo1}. The~main objective of this optimization process is to select the best position candidate for introducing raindrop perturbations that result in minimal classification accuracy (i.e., maximum attack success rate).

Our proposed search method is presented in Algorithm \ref{algo1}: We start by initializing the number of iterations $T$ and the number of candidate positions $N$. We then initialize the perturbation as an empty set.
For each iteration $t$ from 1 to $T$, we randomly generate $N$ candidate positions within the image. For~each candidate position, we apply the Gaussian blur and fish-eye effect to create the raindrop pattern. We simulate the effect of the raindrop perturbation on the image using the candidate position and use Grad-Cam to identify the most critical regions for the model's decision. Then, we select the candidate position that results in the highest attack success rate (i.e., the~lowest classification accuracy). After~that, we update the perturbation with the raindrop pattern from the selected candidate position. We then return the final perturbation, which represents the optimal raindrop pattern for introducing adversarial perturbations with minimal classification~accuracy.

This grad-cam-guided random search-based optimization method efficiently explores different positions for introducing the raindrop perturbations, focusing on regions that significantly influence the model's decision. By~selecting the position that maximizes the attack success rate, we ensure that the adversarial camera sticker is effective in deceiving the target deep learning model while appearing inconspicuous in the captured~images.

\begin{algorithm}[H]
\caption{Grad-CAM-guided Random Search~Method.}
\label{algo1}
\begin{algorithmic}[1]



\State Initialize the number of iterations $T$ and the number of candidate positions $N$
\State Initialize the perturbation as an empty set.
\State Use Grad-Cam to identify the most critical regions for the model's decision.
\For{$t = 1 : T$}
    \State Randomly generate $N$ candidate positions within the identified critical regions.  
    \For {$p = 1 : N$} 
        \State Apply the Gaussian blur and fish-eye effect to create the raindrop pattern.
        \State Simulate the effect of the raindrop perturbation on the image using the candidate position.
        \State Select the candidate position that results in the highest attack success rate.
        \State Update the perturbation with the raindrop pattern from the selected candidate position.
       \EndFor
\EndFor
    
\end{algorithmic}
\end{algorithm}

\subsection{Raindrop~Generator}

The drop generator applies a Gaussian blur transformation to each pixel belonging to the area covered by the raindrop. The~transformation function is defined as follows:
Let $I(i, j)$ be the input image, where $(i, j)$ are the pixel coordinates, and~$I'(i, j)$ be the resulting image after applying the Gaussian blur transformation for the~raindrop.

The Gaussian blur transformation is given by
\begin{equation}
    I'(i, j) = \frac{1}{\sum_{a,b}W(n,m)}\sum_{a,b}I(i+a, j+b)\dot W(a,b)
\end{equation}
where $W(a, b)$ is the Gaussian kernel defined as
\begin{equation}
    W(a,b) = \frac{1}{2 \pi \sigma^2}\exp{-\frac{a^2+b^2}{2\sigma^2}}
\end{equation}
here, 
 $a$ and $b$ are the pixel offsets within the raindrop area, and~$\sigma$ is the standard deviation of the Gaussian kernel, controlling the blur~intensity.

The application of the Gaussian blur transformation ensures that the raindrop pattern appears natural and seamless in the image, emulating the effect of raindrops on the camera lens due to atmospheric conditions. This approach contributes to the inconspicuousness of the adversarial perturbation, making it difficult to detect by human observers while effectively deceiving the target deep learning model.

\begin{figure}[H]
\hspace{-4pt}\includegraphics[width=0.6\columnwidth]{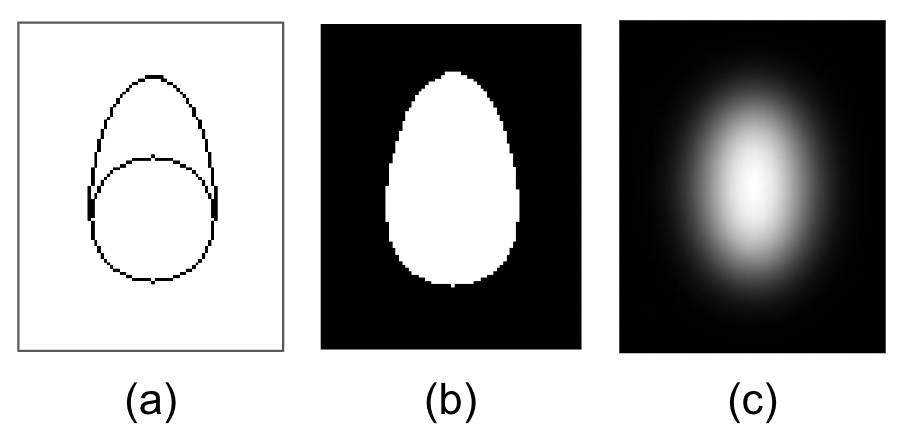}
    \caption{Raindrop generation process: (\textbf{a}) The raindrop shape is formed using one circle and one oval. (\textbf{b}) The final shape. (\textbf{c}) We then create the effect of the water droplet surface through adding the blur~effect.}
\label{drop}
\end{figure}

To create the realistic effect of raindrop surfaces, our approach involves both Gaussian blur and a fish-eye effect. The~process of simulating raindrops and their splashing is achieved through collision detection, where we check if the center of a raindrop overlaps with another raindrop. If~there is an overlap, the~raindrops are merged; otherwise, no action is taken. To~craft the shape of raindrops, we utilize a combination of one circle and one oval (Figure \ref{drop}a). By~manipulating the size and orientation of these shapes, we can create different gaps and simulate the appearance of water droplets (Figure \ref{drop}b). The~final effect of the water droplet surface is achieved through the application of Gaussian blur (Figure \ref{drop}c). The~generated perturbation, representing the raindrop patterns, is then printed on a translucent sticker. This sticker is carefully affixed to the camera lens, allowing the perturbations to be introduced into the visual path without obstructing the image capture~process.

The combination of the fish-eye effect, collision detection, and~raindrop shape design, along with the Gaussian blur, results in realistic and inconspicuous raindrop patterns on the camera lens. These patterns effectively mislead the target deep learning model while resembling a natural weather phenomenon. This approach demonstrates the effectiveness of our proposed AdvRain attack, highlighting its potential impact on camera-based vision systems and the significance of robust defense mechanisms to counter physical adversarial attacks in real-world applications. Figure~\ref{drop} showcases the concept and various stages of raindrop pattern~creation.






\section{Experimental~Results}
\label{sec:experiment}
\unskip

\subsection{Experimental~Setup}
In our study, we conducted experiments to evaluate the impact of different sizes of raindrops on the classification accuracy of VGG-19~\cite{vgg19} and Resnet34~\cite{resnet}, which serve as the victim classifiers. The~input image resolution for both models is set to $224 \times 224$ pixels.

\begin{figure}[H]
\includegraphics[width=0.8\columnwidth]{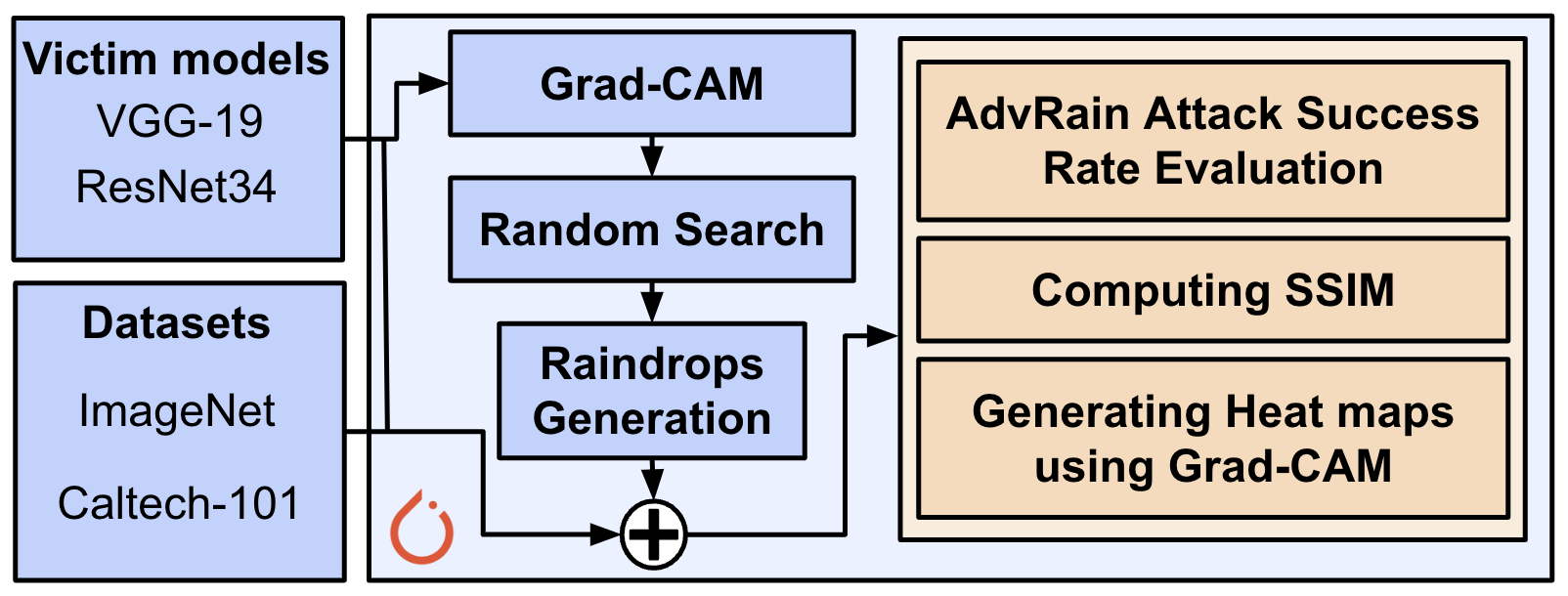}
    \caption{Experimental 
 setup and tool-flow for conducting our~experiments.}
\label{experiments}
\end{figure}

For our evaluations, we utilized images from two well-known datasets: ImageNet and Caltech-101~\cite{caltech}. ImageNet~\cite{imagenet} is a large visual dataset containing over \mbox{14 million} annotated images, organized into more than 20,000 categories. On~the other hand, Caltech-101 consists of images depicting objects from 101 distinct classes, with~approximately 9000 images in total. Each class contains a varying number of images, ranging from 40 to 800, and~the images have variable sizes with typical edge lengths of 200--300 pixels. Throughout our experiments, we established the baseline classification accuracy of the models by evaluating their performance when fed with clean, unaltered images. By~simulating different sizes of raindrops and studying their effects on model classification accuracy, we gained valuable insights into the robustness of the models under adversarial perturbations. The~comparison to the baseline classification accuracy allowed us to quantitatively assess the effectiveness of our AdvRain attack and its ability to deceive the models while introducing minimal visual changes to the images. The experimental setup is summarized in Figure \ref{experiments}.

\subsection{Evaluation of Attack~Performance}
In our evaluation, we adopted classification accuracy as the primary metric to assess the effectiveness of our AdvRain attack. To~demonstrate the impact of the attack, we first generated the adversarial perturbation for 10 different classes and then measured the model accuracy for various numbers of~raindrops.

The results of our experiments revealed that our attack successfully reduced the classification accuracy of the models. Specifically, with~just 10 raindrops applied, we observed a significant decrease in accuracy of approximately $30\%$ for the ImageNet dataset and $20\%$ for the Caltech-101 dataset. As~the number of raindrops increased to 20, the~drop in accuracy became more pronounced, reaching more than $40\%$ for ImageNet and over $30\%$ for Caltech-101 (as depicted in Table~\ref{averageAcc}).
\begin{table}[H]
    \caption{Impact 
of drop radius on model accuracy for VGG-19 trained for ImageNet and Resnet34 trained on Caltech-101: The bigger the radius of the drop the higher drop in classification accuracy we~achieve.}
    \label{averageAcc}
   \setlength{\tabcolsep}{14.5mm}
    \begin{tabularx}{\textwidth}{ccc}
    \toprule
       \textbf{Number of Drops} & \textbf{VGG-19} & \textbf{Resnet34}\\
       \midrule
       \textbf{0 
}    & 100\%  &   100\%\\
        \textbf{10}  &  72\% &   81\%\\
        \textbf{20}  &  59\% &   69\%\\
        \textbf{30}  &  46\% &   50\%\\
        \textbf{40}  &  39\% &   43\%\\
       \bottomrule
    \end{tabularx}
\end{table}

These findings indicate the potency of the AdvRain attack in misclassifying objects of different classes while maintaining inconspicuous patterns resembling natural weather events. The~considerable reduction in model accuracy with just a small number of raindrops highlights the susceptibility of camera-based vision systems to physical adversarial attacks, underscoring the need for robust defense strategies to counter such threats in real-world~applications.

In Table~\ref{perclass}, we present the per-class classification accuracy for eight different classes: Walker hound, Cock, Snake, Spider, Fish, Parrot, American flamingo, and~Bison. Among~these classes, the~''Walker hound'' class exhibited the highest drop in accuracy when subjected to our AdvRain attack. This can be attributed to the fact that the ImageNet dataset includes a total of 120 categories of dog breeds, many of which have close features and similarities. Consequently, the~presence of raindrop perturbations in the images of Walker hounds, which share visual characteristics with other dog breeds, led to a more substantial decrease in classification~accuracy.

\begin{table}[H]
    \caption{Average Model Accuracy per class for different number of raindrops (Drop radius = 10) for VGG-19 on~ImageNet.}
    \label{perclass}
     \setlength{\tabcolsep}{2.9mm}
    \begin{tabularx}{\textwidth}{cccccccccc}
    \toprule
         \textbf{Number}             & \textbf{Walker} & \textbf{} & \textbf{} & \textbf{} & \textbf{} & \textbf{} &  \textbf{American} &\textbf{}\\
        \textbf{of Drops}& \textbf{Hound} & \textbf{Cock} &\textbf{Snake} & \textbf{Spider} & \textbf{Fish} & \textbf{Parrot}  & \textbf{Flamingo} &\textbf{Bison}\\
       \midrule
       10       & 43\%  & 70\%  & 79\% &  68\% & 66\%  &  86\% & 78\% & 68\% \\
       20       & 22\%  & 56\%  & 63\% & 59\%  & 58\% &  82\% & 69\% & 57\%\\
       30       & 12\%  & 48\%  & 49\% & 51\%  & 35\% &  68\% &  60\% & 33\%\\
       40       &  4\% & 40\%  & 37\% &  43\% & 29\% &  60\% &  52\% & 31\%\\
       \bottomrule
    \end{tabularx}
\end{table}


However, for~classes with more distinguishable and unique features, such as the ``Parrot'' class and the ``American flamingo'', we observed a relatively smaller drop in accuracy. Fooling these objects proves to be more challenging due to their distinctive visual attributes, which made it harder for the raindrop perturbations to significantly mislead the~models.

These per-class results provide valuable insights into the impact of our AdvRain attack on different object categories. The~varying degrees of accuracy drop across classes underscore the importance of considering the visual characteristics and complexity of objects when assessing the effectiveness of adversarial attacks. Such knowledge is crucial for understanding the limitations and strengths of the attack and can guide the development of targeted defense strategies for enhancing the robustness of camera-based vision systems in the face of physical adversarial~perturbations.

\subsection{Comparison with State-of-the-Art: AdvRain vs.~FakeWeather}
\begin{figure}[H]
\hspace{-3.5pt}\includegraphics[width=0.7\columnwidth]{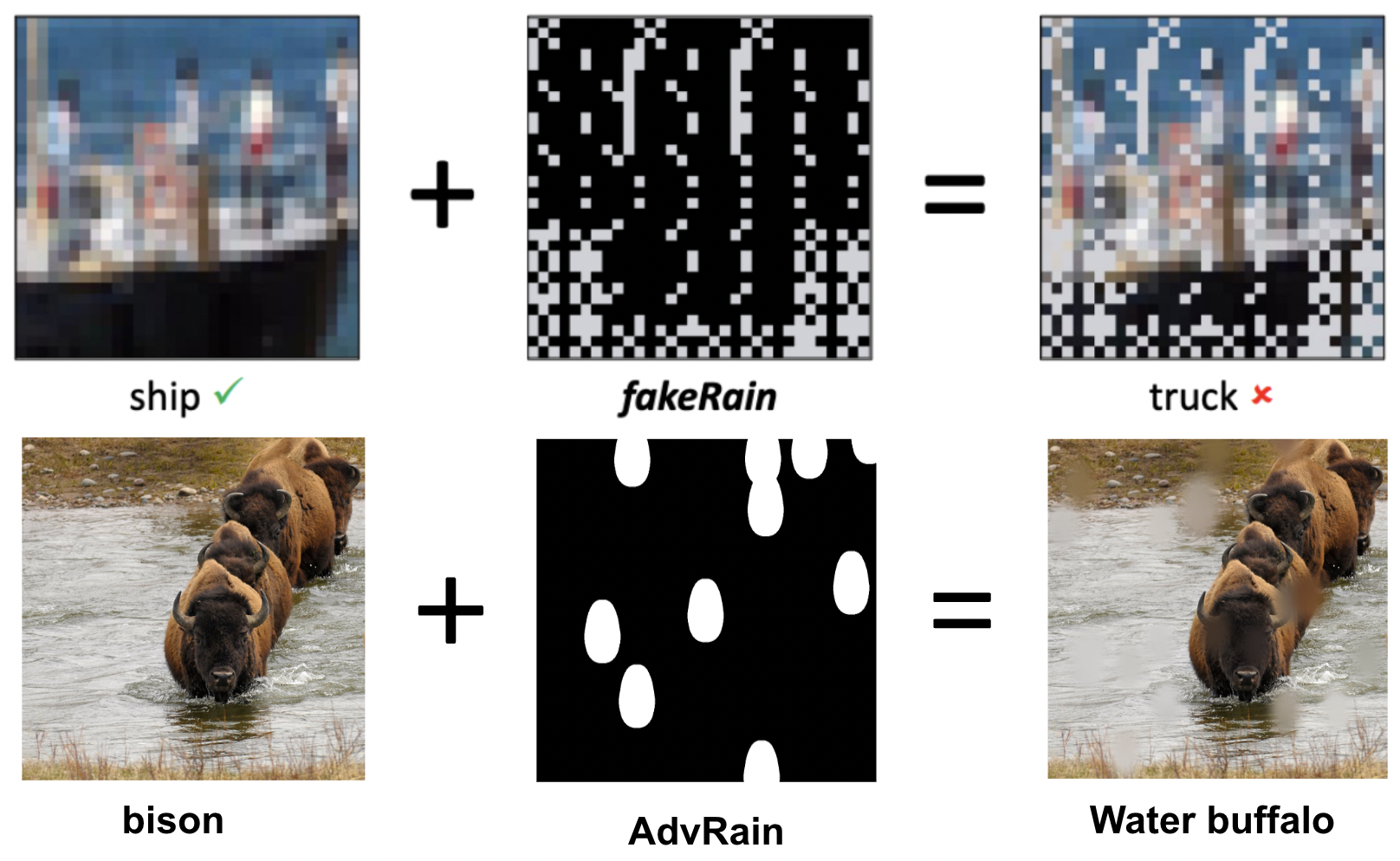}
    \caption{AdvRain compared to FakeWeather attack. FakeWeather attack tries to emulate the rain effect by designing a mask that fakes the effect of such weather conditions on the camera lenses by changing the pixel values. However, the~generated mask resulted in unrealistic and pixelated patterns. In~contrast, AdvRain is based on generating more realistic raindrops simulation with a shape closer to that of a real~raindrop.}
\label{fakeweather}
\end{figure}

In this section, we conduct a comparison between our AdvRain attack and the FakeWeather attack proposed in~\cite{fakeweather}. The~FakeWeather attack aims to simulate the effects of rain, snow, and~hail on camera lenses by creating three masks that mimic the appearance of these weather conditions. However, a~key limitation of the FakeWeather attack is the use of unrealistic and pixelated patterns for generating noise. Additionally, the~added perturbations cover a significant portion of the image (Figure 
 \ref{fakeweather}), potentially making them more conspicuous and easily detectable by human~observers.

Furthermore, the~effectiveness of the FakeWeather attack was evaluated solely on small images with a size of $32\times32$ pixels from the Cifar-10 dataset. In~contrast, our proposed AdvRain attack generates more realistic perturbations that closely emulate the effect of raindrops on camera lenses. The~raindrop patterns crafted by our attack are inconspicuous and blend seamlessly into the image, making them challenging to detect~visually.

Moreover, we extend the evaluation of AdvRain to larger images with a size of $224\times224$ pixels from the ImageNet and Caltech-101 datasets. This broader evaluation showcases the versatility and robustness of our attack on high-resolution images, making it more suitable for real-world scenarios and camera-based vision systems, such as those used in autonomous vehicles and intelligent~robots.

As illustrated in Table~\ref{tab:fake}, our propose attack outperforms FakeWeather when used to attack models trained on ImageNet and Caltech-101. For~instance, AdvRain achieves an attack success rate of 65\%; however, for~the same SSIM, FakeWeather achieves only 37\%.

In summary, our AdvRain attack outperforms the FakeWeather attack by producing more realistic and inconspicuous raindrop perturbations. The~improved emulation of the rain effect and the demonstrated effectiveness on larger images strengthen the applicability and impact of our proposed attack in challenging the integrity of deep learning models in camera-based vision systems. This comparison emphasizes the significance of developing attacks that are not only powerful but also realistic and unobtrusive in real-world~applications.
\unskip

\begin{table}[H]
    \caption{Attack Success Rate: AdvRain vs. FakeWeather~\cite{fakeweather}.}
    \label{tab:fake}
    \setlength{\tabcolsep}{14.5mm}
    \begin{tabularx}{\textwidth}{ccc}
    \toprule
         \textbf{Method} &  \textbf{ImageNet} & \textbf{Caltech-101}\\
         \midrule
        FakeWeather~\cite{fakeweather} & 37\%  & 40\%\\
        AdvRain (ours) & 65\% & 62\%\\
        \bottomrule
    \end{tabularx}
\end{table}
\unskip
\subsection{AdvRain vs. Natural~Rain}

\begin{figure}[H]
\hspace{-8pt}\includegraphics[width=14 cm]{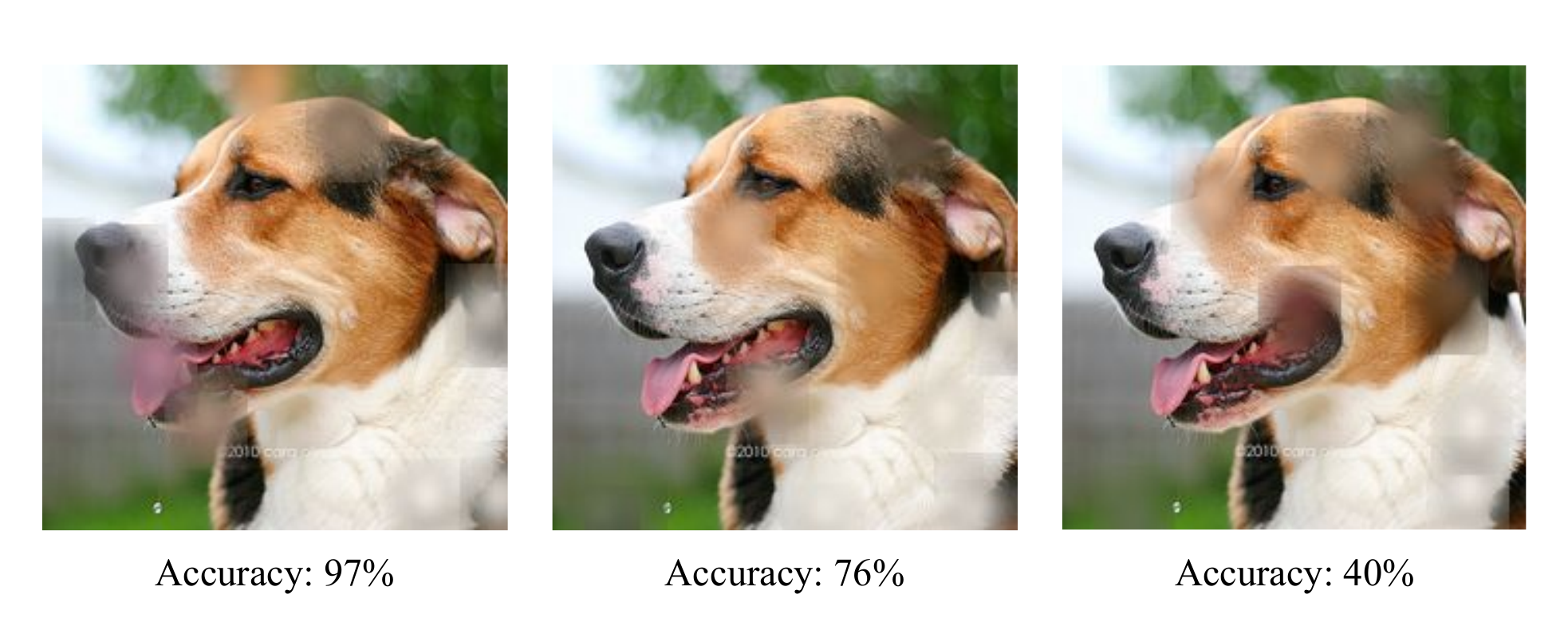}
    \caption{Impact of AdvRain on classification accuracy compared to natural rain. \textit{Left: 
} Natural rain (Random positioning of the raindrops) resulted in a $3\%$ drop. \textit{Middle:} The combination of non-optimal positions resulted in a drop of 23\%. \textit{Right:} the carefully selected combination (AdvRain's output) resulted in a $60\%$ drop.}
\label{naturalvsadv}
\end{figure}

In our evaluation, we conducted a comparison between our AdvRain attack and natural rain. To~simulate natural rain, we randomly placed raindrops in the images, and~this approach resulted in varying degrees of accuracy degradation for the victim model. As~depicted in Figure~\ref{naturalvsadv}, different positions of the randomly placed raindrops led to different accuracy~drops.

The comparison highlights the effectiveness and superiority of our AdvRain attack over natural rain in achieving significant classification accuracy drops. For the left combination of totally random raindrops, the~accuracy drop was only $3\%$, indicating that the random placement of raindrops had limited impact on fooling the victim model. However, when we carefully selected the positions of raindrops using our AdvRain attack, the~accuracy drop reached a significant value of $60\%$.

These results demonstrate that the adversarial perturbations crafted by our AdvRain attack are much more potent in deceiving the deep learning model compared to the impact of natural rain. The~strategic placement of raindrops in AdvRain leads to a higher level of misclassification, indicating that our attack successfully leverages the characteristics of raindrop patterns to create more effective adversarial~perturbations.

This comparison emphasizes the unique advantage of AdvRain in generating inconspicuous yet powerful adversarial perturbations that effectively mislead the target model, surpassing the impact of random raindrop placements associated with natural rain. The~ability to achieve substantial accuracy drops with carefully selected raindrop positions highlights the potential real-world implications of AdvRain, particularly in scenarios involving camera-based vision systems such as those used in autonomous vehicles and intelligent~robots.
\vspace{-8pt}

\section{Discussion}
\label{sec:discussion}
\unskip
\subsection{Evaluation on More DNN~Models}
\textcolor{black}{
AdvRain attack has been tested on other convolutional neural network (CNN) models, specifically Resnet-50 and Inception v4. As~shown in Table~\ref{other_models}, when AdvRain is applied to Resnet-50, it successfully causes misclassification in the target model in approximately 59\% of the cases. In~other words, when images perturbed by AdvRain are fed into Resnet-50, the~model's accuracy drops by 59\% compared to its accuracy on clean, unperturbed images. This demonstrates the effectiveness of the AdvRain attack on Resnet-50. Additionaly, when AdvRain is applied to the Inception v4 model, it achieves a success rate of approximately 57\%. This indicates that the attack causes a 57\% drop in accuracy when applied to Inception v4. This highlights the performance of the AdvRain attack on different CNN models, specifically Resnet-50 and Inception v4. It shows the effectiveness of the attack, which represents the extent to which AdvRain can deceive or mislead these models when applied to images.}

\begin{table}[H]
    \caption{Performance of AdvRain on More CNN~models.}
    \label{other_models}
     \setlength{\tabcolsep}{10mm}
    \begin{tabularx}{\textwidth}{cccccc}
    \toprule
       \textbf{Model}  & \textbf{VGG-19} & \textbf{Resnet-50} & \textbf{Inception v4} \\
       \midrule
       Accuracy   & 37\%  & 41\% & 43\% \\
       \bottomrule
    \end{tabularx}
\end{table}
\unskip
\subsection{Impact of Drop~Radius}
\textcolor{black}{
In our evaluation, we systematically investigated the impact of changing the radius of the raindrop (i.e., the~blurred area) on the effectiveness of our AdvRain attack. We tested three different models trained on ImageNet (i.e., VGG-19, Resnet-50, Inception v4) and three others trained on Caltech-101 (i.e., Resnet-34, Resnet-50, and~Inception v4).
We observed that increasing the radius led to a substantial increase in the attack success rate, further compromising the accuracy of the victim model. For~instance, when we increased the radius from $10$ to $15$ pixels for the ImageNet dataset, the~accuracy of the victim model dropped an additional $25\%$ (as depicted in Tables~\ref{impact_radius_imagenet} and \ref{impact_radius_caltech}). This finding demonstrates the significance of the raindrop size in determining the strength of the adversarial perturbations.}

\begin{table}[H]
    \caption{Impact of drop radius on model accuracy for ImageNet: The bigger the radius of the drop, the higher drop in classification accuracy we~achieve.}
    \label{impact_radius_imagenet}
   \setlength{\tabcolsep}{7mm}
    \begin{tabularx}{\textwidth}{cccccc}
    \toprule
       \textbf{Number of Drops}  & \textbf{0} & \multicolumn{2}{c}{\textbf{10}} & \multicolumn{2}{c}{\textbf{20}} \\
       \midrule
       Radius         & -  & 10 & 15 &10 & 15 \\
       \midrule
       VGG-19       & 100\%  & 76\%  & 57\% & 59\%  & 37\% \\
       \bottomrule
    \end{tabularx}
\end{table}

The larger the radius of the raindrop, the~more extensive the area covered by the perturbation, making it more impactful in fooling the deep learning model. The~increased attack success rate with larger raindrop sizes indicates that our AdvRain attack is particularly effective in~situations where the raindrop pattern covers a significant portion of the image, emulating the effect of real-world raindrops on the camera~lens.

\begin{table}[H]
    \caption{Impact of drop radius on model accuracy for Caltech-101: The bigger the radius of the drop the higher drop in classification accuracy we~achieve.}
    \setlength{\tabcolsep}{7mm}
    \begin{tabularx}{\textwidth}{cccccc}
    \toprule
       \textbf{Number of Drops} & \textbf{0} & \multicolumn{2}{c}{\textbf{10}} & \multicolumn{2}{c}{\textbf{20}} \\
       \midrule
       Radius         & -  & 10 & 15 &10 & 15 \\
       \midrule
       Resnet-34      & 100\%   & 79\%  & 70\% & 69\%  & 55\% \\
       \bottomrule
    \end{tabularx}
    \label{impact_radius_caltech}
\end{table}

\subsection{Impact on~SSIM}

In this section, we focus on evaluating the impact of the adversarial perturbation on the captured images using the structural similarity index measure (SSIM) \cite{ssim}. Given that our AdvRain attack is designed to introduce raindrop perturbations by blurring specific regions of the image, we observe a relatively small impact on SSIM when compared to traditional adversarial attacks (Table \ref{ssim}). For~instance, when using 10 raindrops, the~SSIM of the adversarial example compared to the clean image is measured at $0.89$. This indicates that the introduction of raindrop perturbations has a limited effect on the structural similarity between the adversarial and clean~images.

The relatively high SSIM values suggest that our AdvRain attack achieves its goal of creating inconspicuous perturbations that closely resemble natural weather phenomena. Despite causing a significant drop in classification accuracy, the~raindrop perturbations retain a high degree of structural similarity to the original image, making them difficult for human observers to detect visually. This aspect sets our AdvRain attack apart from traditional adversarial attacks that often introduce noticeable and disruptive patterns in the image, resulting in lower SSIM values. The~inconspicuous nature of our perturbations contributes to the stealthiness and effectiveness of AdvRain in deceiving camera-based vision systems without significantly altering the visual appearance of the captured~images.
\begin{table}[H]
\centering
  \caption{SSIM of adversarial examples compared to the original~images.}
  \label{ssim}
  \setlength{\tabcolsep}{9.5mm}
    \begin{tabularx}{\textwidth}{ccccc}
    \toprule
       \textbf{Number 
 of Drops}   & \textbf{10} & \textbf{20}  & \textbf{30} & \textbf{40} \\
    \midrule 
        SSIM       &  0.89  &  0.78 & 0.73 & 0.69 \\

  \bottomrule
\end{tabularx} 
\end{table}
\unskip

\subsection{Impact on Network~Interpretation }

Adversarial patches have demonstrated their efficacy in causing misclassification, but~their usage has also raised concerns about their detectability. Standard network interpretation methods, such as Grad-CAM, can highlight the presence of adversarial patches, effectively disclosing the identity of the adversary~\cite{grad}. Grad-CAM, being one of the most well-known network interpretation algorithms, has proven to outperform other state-of-the-art interpretation methods in various~scenarios.

To evaluate the Grad-CAM visualization results for traditional adversarial patches versus our AdvRain attack, we utilized an ImageNet pre-trained VGG-19 classifier. The~evaluation involved comparing the effects of adding low frequency patterns (traditional adversarial patch) versus high frequency patterns (AdvRain) on the model's~interpretation.

\begin{figure}[H]
\hspace{-14pt}\includegraphics[width=\textwidth]{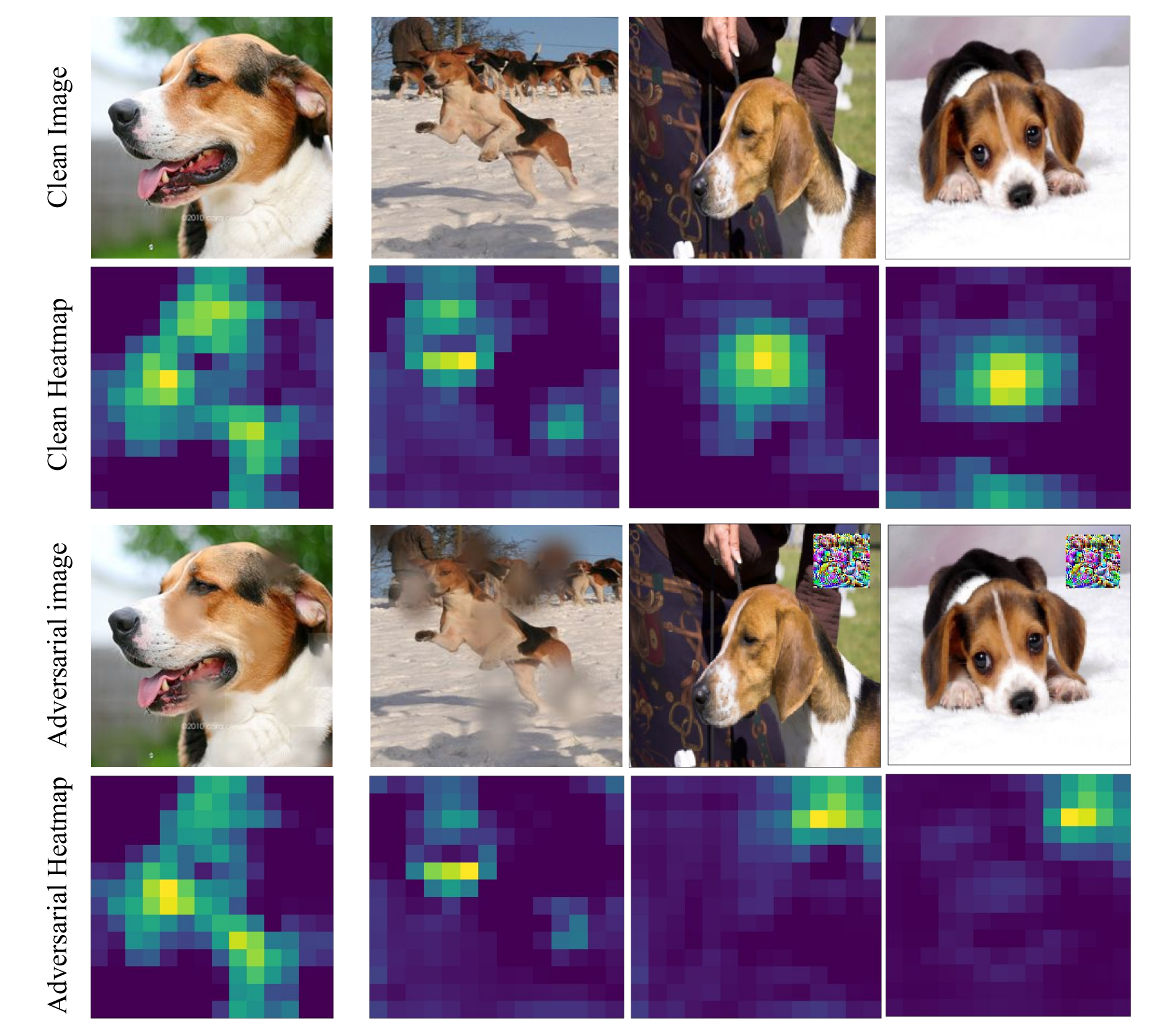}
    \caption{Comparing 
 the Grad-CAM visualization results for adversarial patch vs.~AdvRain.}
\label{gradcam}
\end{figure}

Unlike patch-based attacks that shift the model's focus from the object to the location of the patch, potentially making them detectable, our AdvRain attack exhibits a different behavior. It causes the model to overlook some crucial features that are essential for the model to make accurate decisions (as illustrated in Figure~\ref{gradcam}). By~strategically introducing raindrop perturbations, we divert the model's attention away from vital features, making it more susceptible to~misclassification.

This behavior reinforces the stealthiness of our AdvRain attack since it does not draw attention to a specific patch or region, unlike traditional adversarial patches. The~inconspicuous nature of the raindrop perturbations, combined with their impact on the model's interpretation, highlights the effectiveness of AdvRain in deceiving camera-based vision systems without raising suspicions about the presence of an adversarial~attack.

The Grad-CAM visualization results provide further evidence of the unique advantages of our AdvRain attack, emphasizing its potential as a powerful and inconspicuous technique for crafting adversarial perturbations that elude standard detection methods and undermine the reliability of deep learning models in real-world applications.

\subsection{Possible~Defenses}
\textcolor{black}{
One potential defense strategy against AdvRain involves training a denoising/deraining model. This approach aims to enhance the resilience of the DNN against adversarial attacks of this nature. The~denoiser operates as a preprocessing stage, effectively eliminating the adversarial noise from the input data before it reaches the target model. This denoising model is specifically designed to mitigate the impact of AdvRain perturbations. }

\textcolor{black}{
The denoising/deraining model can be designed to learn how to cancel or neutralize the influence of AdvRain perturbations on the model's output. This can be achieved by defining a loss function that guides the training of the denoising model. Here is how this process can work: The loss function can be defined to measure the dissimilarity between the predictions made by the victim model when given a clean image and when given a denoised (derained) image. The~goal is to minimize this dissimilarity, effectively forcing the denoising model to remove AdvRain perturbations in a way that ensures the victim model's output remains consistent with that of a clean image. The~training data will consist of pairs of images: Clean Images (i.e., images that have not been subjected to AdvRain attacks) and AdvRain-Infected Images (i.e., images that have been attacked by AdvRain to introduce perturbations).}

\textcolor{black}{
The denoising model is trained on these pairs of images to minimize the defined loss function. During~training, the~model learns to identify and understand the patterns associated with AdvRain perturbations. It also learns how to modify the AdvRain-infected image to cancel out these perturbations, effectively generating a denoised version that closely resembles the clean image. After~training, when the denoising model receives an AdvRain-infected image as input, it applies the learned transformations to cancel out the AdvRain perturbations. The~derained image, which is free of AdvRain noise, is then passed to the victim model for classification or other tasks. Because~the denoising model has effectively removed the AdvRain influence, the~victim model's output remains consistent with what it would produce when given a clean image. This approach enhances the robustness of the system against AdvRain attacks. The~denoising model acts as a filter that ``cleans'' AdvRain-infected images before they reach the victim model, ensuring that the model's predictions are not adversely affected by the perturbations.}

\textcolor{black}{
Another defense strategy can be to generate adversarial samples considering different rain models, and~incorporate these samples during the adversarial training process for a given DNN. 
By training the DNN on this augmented dataset that encompasses a variety of AdvRain-induced adversarial samples, the~model can become more robust and resilient to AdvRain attacks across a wider range scenarios. }
\subsection{AdvRain on~Videos}
\textcolor{black}{
One notable feature of AdvRain is its ability to conduct class-wide attacks. This means that the same adversarial perturbation, represented by the raindrop pattern on the sticker attached to the camera of a system (e.g., cars), can be universally applied to different images of objects within the same class. This universal approach simplifies the attack process and efficiently deceives the target model across a broad range of objects within the targeted class. Regarding the application of AdvRain in videos, its effectiveness depends on several factors, with~context stability being a key consideration. We believe that AdvRain can maintain its effectiveness in video scenarios where the context remains relatively stable. In~such cases, the~attack can continue to deceive the model effectively in video scenario. This phenomenon is commonly observed in video scenes characterized by consistent visual elements.}

\subsection{Limitations}
\textcolor{black}{
AdvRain is an effective and stealthy approach to adversarial attacks, but~like any technique, it may have some potential limitations, as~discussed~below: 
\begin{itemize}
    \item \textbf{Device-Specific}
: The success of AdvRain can depend on the specific camera and lens characteristics of the target device. Variations in camera types, lens coatings, and~sensor resolutions can affect the effectiveness of the attack. This means that the same AdvRain sticker may not work with same efficiency on all devices.
    \item \textbf{Sticker Placement}: The success of the attack depends on the precise placement of the AdvRain sticker on the camera lens. If~the sticker is misaligned or partially obstructed, it may not create the desired perturbations, reducing the attack's effectiveness.
    \item \textbf{Environmental Conditions}: The effectiveness of AdvRain can be influenced by environmental conditions such as lighting, weather, and~visibility. Raindrop patterns may be more or less convincing under different conditions, potentially limiting the attack's reliability.
\end{itemize}
}



\section{Conclusions}
\label{sec:conclusion}

In this paper, we present a novel approach for crafting realistic physical adversarial camera stickers to deceive image classifiers. Our proposed attack, known as AdvRain, emulates the effect of natural weather conditions, specifically raindrops, when placed on the camera lens. Our methodology revolves around utilizing random search-based optimization methods to identify the optimal raindrop positions. Through extensive evaluations, we demonstrate the effectiveness of AdvRain. With~just 20 raindrops, we achieve a significant drop in average classification accuracy, exceeding 45\% and 40\% for VGG19 on ImageNet and Resnet34 on Caltech-101, respectively. Our results showcase the potency of AdvRain as a stealthy and powerful approach to adversarial attacks in camera-based vision systems.








\pagebreak
\authorcontributions{Conceptualization, A.G., M.A.H. and M.S.; methodology, A.G., M.A.H. and M.S.; software, A.G.; validation, A.G.; formal analysis, A.G.; investigation, A.G.; resources, A.G.; data curation, A.G.; writing---original draft preparation, A.G., M.A.H. and M.S.; writing---review and editing, M.S.; visualization, A.G.; supervision, M.S.; project administration, M.S.; funding acquisition, M.S. All authors have read and agreed to the published version of the manuscript.
}

\funding{
This work was supported in parts by the NYUAD Center for Cyber Security (CCS), funded by Tamkeen under the NYUAD Research Institute Award G1104, Center for Interacting Urban Networks (CITIES), funded by Tamkeen under the NYUAD Research Institute Award CG001, and~Center for Artificial Intelligence and Robotics (CAIR), funded by Tamkeen under the NYUAD Research Institute Award CG010.
}

\dataavailability{No new data were created.} 

\conflictsofinterest{The authors declare no conflict of~interest.} 

\begin{adjustwidth}{-\extralength}{0cm}

\reftitle{References}

\PublishersNote{}
\end{adjustwidth}

\begin{thebibliography}{999}

\bibitem[Al-Qizwini, Mohammed ; Barjasteh, Iman ; Al-Qassab, Hothaifa ; Radha, Hayder]{al2017deep}
\newblock Al-Qizwini, M. ; Barjasteh, I. ; Al-Qassab, H. ; Radha, H. 
\newblock Deep learning algorithm for autonomous driving using GoogLeNet. \emph{IEEE Intelligent Vehicles Symposium (IV)} \textbf{2017}, 89--96.

\bibitem[Tram{\`e}r et~al.(2016)Tram{\`e}r, Zhang, Juels, Reiter, and
  Ristenpart]{b0}
Tram{\`e}r, F.; Zhang, F.; Juels, A.; Reiter, M.K.; Ristenpart, T.
\newblock Stealing Machine Learning Models via Prediction APIs.
\newblock In Proceedings of the 25th {USENIX} Security Symposium ({USENIX}
  Security 16), Austin, TX, USA, 10--12 August 2016; pp. 601--618.

\bibitem[Goodfellow et~al.(2015)Goodfellow, Shlens, and Szegedy]{fgsm}
Goodfellow, I.J.; Shlens, J.; Szegedy, C.
\newblock Explaining and Harnessing Adversarial Examples.  \emph{arXiv} \textbf{2015},  arXiv:1412.6572. 

\bibitem[Kurakin et~al.(2016)Kurakin, Goodfellow, and Bengio]{phy9}
Kurakin, A.; Goodfellow, I.J.; Bengio, S.
 Adversarial examples in the physical world.
\newblock {\em arXiv} {\bf 2016}, arXiv:1607.02533.


\bibitem[Carlini and Wagner(2016)]{CW}
Carlini, N.; Wagner, D.A.
\newblock Towards Evaluating the Robustness of Neural Networks.
\newblock {\em arXiv} {\bf 2016}, arXiv:1608.04644.

\bibitem[Zhong et~al.(2022)Zhong, Liu, Zhai, Jiang, and Ji]{Zhong2022}
Zhong, Y.; Liu, X.; Zhai, D.; Jiang, J.; Ji, X.
\newblock Shadows can be Dangerous: Stealthy and Effective Physical-world
  Adversarial Attack by Natural Phenomenon. {\em arXiv} {\bf 2022},  arXiv:2209.02430.
\newblock {\url{https://doi.org/10.48550/ARXIV.2203.03818}}.

\bibitem[Hu and Shi(2022)]{Hu2022}
Hu, C.; Shi, W.
\newblock Adversarial Color Film: Effective Physical-World Attack to DNNs. 
  {\em arXiv} {\bf 2022}, arXiv:2209.02430.
\newblock {\url{https://doi.org/10.48550/ARXIV.2209.02430}}.

\bibitem[Duan et~al.(2020)Duan, Ma, Wang, Bailey, Qin, and Yang]{Duan2020}
Duan, R.; Ma, X.; Wang, Y.; Bailey, J.; Qin, A.K.; Yang, Y.
\newblock Adversarial Camouflage: Hiding Physical-World Attacks with Natural
  Styles. {\em arXiv} {\bf 2020},  	arXiv:2003.08757.
\newblock {\url{https://doi.org/10.48550/ARXIV.2003.08757}}.

\bibitem[Marchisio et~al.(2022)Marchisio, Caramia, Martina, and
  Shafique]{fakeweather}
Marchisio, A.; 
 Caramia, G.; Martina, M.; Shafique, M.
\newblock fakeWeather: Adversarial Attacks for Deep Neural Networks Emulating
  Weather Conditions on the Camera Lens of Autonomous Systems. \emph{arXiv} \textbf{2022}, arXiv:2205.13807.
\newblock {\url{https://doi.org/10.48550/ARXIV.2205.13807}}.

\bibitem[Brown et~al.(2017)Brown, Mane, Roy, Abadi, and Gilmer]{googleap}
Brown, T.; Mane, D.; Roy, A.; Abadi, M.; Gilmer, J.
\newblock Adversarial Patch.
\newblock  {\em arXiv} {\bf 2017}, arXiv:1712.09665.

\bibitem[Subramanya et~al.(2018)Subramanya, Pillai, and Pirsiavash]{grad}
Subramanya, A.; Pillai, V.; Pirsiavash, H.
\newblock Towards Hiding Adversarial Examples from Network Interpretation.
\newblock {\em arXiv} {\bf 2018}, arXiv:1812.02843.

\bibitem[Szegedy et~al.(2014)Szegedy, Zaremba, Sutskever, Bruna, Erhan,
  Goodfellow, and Fergus]{Szegedy}
Szegedy, C.; Zaremba, W.; Sutskever, I.; Bruna, J.; Erhan, D.; Goodfellow,
  I.J.; Fergus, R.
\newblock Intriguing properties of neural networks.
\newblock In \emph{Proceedings of the 2nd International Conference on Learning
  Representations, {ICLR} 2014, Banff, AB, Canada, 14--16 April  2014}; Conference Track Proceedings; Bengio, Y., LeCun, Y., Eds.;   2014. 


\bibitem[Brendel et~al.(2017)Brendel, Rauber, and Bethge]{BA}
Brendel, W.; Rauber, J.; Bethge, M.
\newblock Decision-Based Adversarial Attacks: Reliable Attacks Against
  Black-Box Machine Learning Models.  {\em arXiv} {\bf 2017}, arXiv:1712.04248.

\bibitem[Narodytska and Kasiviswanathan(2016)]{localsearch}
Narodytska, N.; Kasiviswanathan, S.P.
\newblock Simple Black-Box Adversarial Perturbations for Deep Networks.
\newblock  {\em arXiv} {\bf 2016}, arXiv:1612.06299.

\bibitem[Chen and Jordan(2019)]{HSJ}
Chen, J.; Jordan, M.I.
\newblock Boundary Attack++: Query-Efficient Decision-Based Adversarial Attack.
\newblock  {\em arXiv} {\bf 2019}, arXiv:1904.02144.

\bibitem[Hu et~al.(2021)Hu, Chen, Kung, Hua, and Tan]{Hu21}
Hu, Y.C.T.; Chen, J.C.; Kung, B.H.; Hua, K.L.; Tan, D.S.
\newblock Naturalistic Physical Adversarial Patch for Object Detectors.
\newblock In Proceedings of the 2021 IEEE/CVF International Conference on
  Computer Vision (ICCV),  Montreal, QC, Canad, 10--17 October 2021; pp. 7828--7837.
\newblock {\url{https://doi.org/10.1109/ICCV48922.2021.00775}}.

\bibitem[Guesmi et~al.(2023{\natexlab{a}})Guesmi, Bilasco, Shafique, and
  Alouani]{guesmi2023advart}
Guesmi, A.; Bilasco, I.M.; Shafique, M.; Alouani, I.
\newblock AdvART: Adversarial Art for Camouflaged Object Detection Attacks.
\newblock {\em arXiv} {\bf 2023}, arXiv:2303.01734.

\bibitem[Guesmi et~al.(2023{\natexlab{b}})Guesmi, Ding, Hanif, Alouani, and
  Shafique]{guesmi2023dap}
Guesmi, A.; Ding, R.; Hanif, M.A.; Alouani, I.; Shafique, M.
\newblock DAP: A Dynamic Adversarial Patch for Evading Person Detectors.
\newblock {\em arXiv} {\bf 2023}, arXiv:2305.11618.

\bibitem[Athalye et~al.(2018)Athalye, Engstrom, Ilyas, and Kwok]{eot}
Athalye, A.; Engstrom, L.; Ilyas, A.; Kwok, K.
\newblock Synthesizing robust adversarial examples.
\newblock \emph{PMLR}  \textbf{2018}, \emph{80}, 284--293. 

\bibitem[Sayles et~al.(2020)Sayles, Hooda, Gupta, Chatterjee, and
  Fernandes]{rolling}
Sayles, A.; Hooda, A.; Gupta, M.; Chatterjee, R.; Fernandes, E.
\newblock Invisible Perturbations: Physical Adversarial Examples Exploiting the
  Rolling Shutter Effect.
\newblock {\em arXiv} {\bf 2020}, arXiv:2011.13375.

\bibitem[Kim et~al.(2021)Kim, Kim, Song, Choi, Joo, and Lee]{slmattack}
Kim, K.; Kim, J.; Song, S.; Choi, J.H.; Joo, C.; Lee, J.S.
\newblock Light Lies: Optical Adversarial Attack. {\em arXiv} {\bf 2021}, arXiv:2106.09908.

\bibitem[Gnanasambandam et~al.(2021)Gnanasambandam, Sherman, and Chan]{optical}
Gnanasambandam, A.; Sherman, A.M.; Chan, S.H.
\newblock Optical Adversarial Attack.
\newblock {\em arXiv} {\bf 2021}, arXiv:2108.06247.

\bibitem[Kennedy and Eberhart(1995)]{488968}
Kennedy, J.; Eberhart, R.
\newblock Particle swarm optimization.
\newblock In Proceedings of the Proceedings of ICNN'95---International
  Conference on Neural Networks, Perth, Australia, 27 November--1 December 1995; Volume 4, pp. 1942--1948.
\newblock {\url{https://doi.org/10.1109/ICNN.1995.488968}}.

\bibitem[Kingma and Ba(2014)]{adam}
Kingma, D.P.; Ba, J.
\newblock Adam: A Method for Stochastic Optimization. {\em arXiv} {\bf 2014},  	arXiv:1412.6980.
\newblock {\url{https://doi.org/10.48550/ARXIV.1412.6980}}.

\bibitem[Simonyan and Zisserman(2014)]{vgg19}
Simonyan, K.; Zisserman, A.
\newblock Very deep convolutional networks for large-scale image recognition.
\newblock {\em arXiv} {\bf 2014}, arXiv:1409.1556.

\bibitem[He et~al.(2015)He, Zhang, Ren, and Sun]{resnet}
He, K.; Zhang, X.; Ren, S.; Sun, J.
\newblock Deep Residual Learning for Image Recognition.
\newblock {\em arXiv} {\bf 215}, arXiv:1512.03385.

\bibitem[Li et~al.(2022)Li, Andreeto, Ranzato, and Perona]{caltech}
Li, F.F.; Andreeto, M.; Ranzato, M.; Perona, P.
\newblock \emph{Caltech 101}. 
\newblock CaltechDATA 
2022. 
\newblock {\url{https://doi.org/10.22002/D1.20086}}.

\bibitem[Deng et~al.(2009)Deng, Dong, Socher, Li, Li, and Fei-Fei]{imagenet}
Deng, J.; Dong, W.; Socher, R.; Li, L.J.; Li, K.; Fei-Fei, L.
\newblock Imagenet: A large-scale hierarchical image database.
\newblock In Proceedings of the 2009 IEEE Conference on Computer Vision and
  Pattern Recognition, Miami, FL, USA, 20--25 June 2009; pp. 248--255. 


\bibitem[Wang et~al.(2004)Wang, Bovik, Sheikh, and Simoncelli]{ssim}
Wang, Z.; Bovik, A.; Sheikh, H.; Simoncelli, E.
\newblock Image quality assessment: From error visibility to structural
  similarity.
\newblock {\em IEEE Trans. Image Process.} {\bf 2004}, {\em
  13},~600--612.
\newblock {\url{https://doi.org/10.1109/TIP.2003.819861}}.

\end{thebibliography}
\end{document}